\documentclass[10pt,twocolumn,letterpaper]{article}

\usepackage{cvpr}              

\definecolor{cvprblue}{rgb}{0.21,0.49,0.74}
\usepackage[pagebackref,breaklinks,colorlinks,allcolors=cvprblue]{hyperref}
\usepackage{multirow} 
\usepackage{threeparttable}
\usepackage{graphicx}
\usepackage{algorithm}
\usepackage{algorithmic}
\usepackage{amssymb, amsmath}
\usepackage{newfloat}
\usepackage{listings}
\usepackage{siunitx} 
\usepackage{caption}

\def\paperID{} 
\def\confName{CVPR}
\def\confYear{2025}

\title{Unveil Inversion and Invariance in Flow Transformer for Versatile Image Editing}

\author{%
    Pengcheng Xu\textsuperscript{\rm 1,2} \hspace{.1in} 
    Boyuan Jiang\textsuperscript{\rm 2} \hspace{.1in} 
    Xiaobin Hu\textsuperscript{\rm 2}\footnotemark[2] \hspace{.1in} 
    Donghao Luo\textsuperscript{\rm 2} \hspace{.1in} 
    Qingdong He\textsuperscript{\rm 2} \\
    Jiangning Zhang\textsuperscript{\rm 2} \hspace{.1in} 
    Chengjie Wang\textsuperscript{\rm 2} \hspace{.1in} 
    Yunsheng Wu\textsuperscript{\rm 2} \hspace{.1in} 
    Charles Ling\textsuperscript{\rm 1} \hspace{.1in} 
    Boyu Wang\textsuperscript{\rm 1}\footnotemark[1]  \vspace{.2cm} \\
    \textsuperscript{\rm 1}Western University  \hspace{.2in} 
    \textsuperscript{\rm 2}Tencent \\
}

\begin{document}

\twocolumn[{%
\renewcommand\twocolumn[1][]{#1}%
\maketitle
\vspace{-37pt}
\begin{center}
    \centering
    \captionsetup{type=figure}
\includegraphics[width=0.92\textwidth]{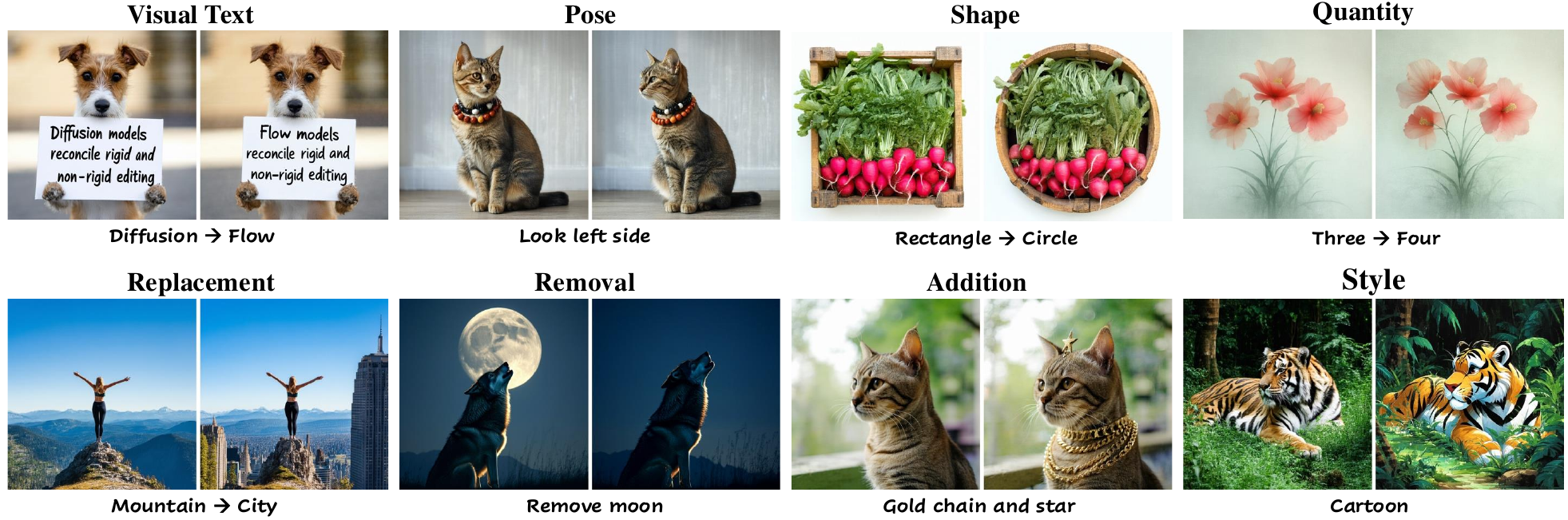}
    \vspace{-10pt}
    \captionof{figure}{\small Our framework reconciles the invariance control for rigid and non-rigid editing, enabling versatile editing via flow transformer.
    }
    \label{fig:teaser}
    \vspace{-10pt}
\end{center}%
}]

\let\thefootnote\relax\footnotetext{$^*$Corresponding author. \quad $^{\dagger}$Project lead.}

\begin{abstract}
Leveraging the large generative prior of the flow transformer for tuning-free image editing requires authentic inversion to project the image into the model's domain and a flexible invariance control mechanism to preserve non-target contents. However, the prevailing diffusion inversion performs deficiently in flow-based models, and the invariance control cannot reconcile diverse rigid and non-rigid editing tasks. To address these, we systematically analyze the \textbf{inversion and invariance} control based on the flow transformer. Specifically, we unveil that the Euler inversion shares a similar structure to DDIM yet is more susceptible to the approximation error. Thus, we propose a two-stage inversion to first refine the velocity estimation and then compensate for the leftover error, which pivots closely to the model prior and benefits editing. Meanwhile, we propose the invariance control that manipulates the text features within the adaptive layer normalization, connecting the changes in the text prompt to image semantics. This mechanism can simultaneously preserve the non-target contents while allowing rigid and non-rigid manipulation, enabling a wide range of editing types such as visual text, quantity, facial expression, etc. Experiments on versatile scenarios validate that our framework achieves flexible and accurate editing, unlocking the potential of the flow transformer for versatile image editing. Project Page is \href{https://pengchengpcx.github.io/EditFT/}{here}.
\end{abstract}    
\section{Introduction}
The success of the diffusion model (DM) in large-scale text-to-image generation~\cite{rombach2022high,podell2023sdxl,saharia2205photorealistic} enables editing images with flexible text-guided interface and rich expressiveness~\cite{hertz2022prompt,kawar2023imagic,parmar2023zero,mokady2023null}. Image editing requires the model conditioning on the input image. The instruction-based methods retrain the text-to-image (T2I) DM by the triplet data of the original-and-edited images, and the editing instruction~\cite{brooks2023instructpix2pix,geng2024instructdiffusion}. However, manually crafting such triples is not scalable and may cause bias~\cite{zhang2024magicbrush}. In contrast, tuning-free methods leverage the model itself to implicitly construct the input image condition, which eliminates additional training and ensures scalability with the large T2I model. This approach has recently garnered significant research interest. Despite this, most studies focus on DM editing, while the potential of recent flow-based T2I models (FMs), which incorporate transformer architecture (i.e., DiT~\cite{peebles2023scalable}), richer data priors, and better text-to-image alignment~\cite{esser2024scaling}, remains underexplored.

In this paper, we investigate tuning-free image editing within the context of the flow transformer from two key aspects: \textbf{inversion} and \textbf{invariance control}. The inversion process projects the image into the model's domain as an initial noisy latent, while invariance control determines which contents from the original image should be preserved during the regeneration of the edited image. The principle of inversion aims to enhance the faithfulness of the recovered original image and editing ability. Although inversions based on the DDIM~\cite{song2020denoising} in DM have been extensively studied, they face significant challenges with the Euler sampler in flow models (FM). Regarding the invariance control, the attention manipulation techniques~\cite{tumanyan2023plug,parmar2023zero,hertz2022prompt} in U-Net dominate DM but are deficient at reconciling rigid and non-rigid editing such as altering numbers, movements, and shapes~\cite{koo2024flexiedit,xu2023inversion,cao2023masactrl}. For instance, preserving non-target contents can hinder changes to the layout and pose of objects. Complementary solutions often require additional diffusion branches and specific refinement~\cite{brack2023sega,koo2024flexiedit}. Thus, a reconciled and efficient solution is beneficial for editing.

Based on the above, the editing challenges of the flow transformer stem from two major issues: \textbf{inversion for flow} and \textbf{invariance control based on the flow transformer}. First, the off-the-shelf diffusion inversions are not optimized for the Euler sampler in flow matching and suffer fallback in practice (Sec:\ref{inv:con}). Developing faithful and editable inversion for flow matching is essential.
Second, how to preserve the invariance (i.e. the unedited contents) based on transformer architecture is crucial for tuning-free editing yet underexplored due to the network architecture difference (DiT v.s. U-Net).
Specifically, MM-DiT~\cite{esser2024scaling} does not have cross-attention as the conditioning mechanism, and simply extending self-attention methods to MM-DiT yields unsatisfied editing. Therefore, effective invariance control must reconcile rigid and non-rigid editing to take full advantage of the rich text-to-image priors for versatile editing. 

To address the aforementioned issues, we systematically investigate flow inversion with the Euler method and invariance modeling in MM-DiT. For inversion, we reveal that the reverse ordinary differential equation (ODE) with the Euler method has a similar structure to DDIM inversion but is more affected by the approximation error between two consecutive states. This error makes Euler inversion diverge more from the original image than DDIM inversion in diffusion and significantly deteriorates the editing ability even if the image can be recovered by optimization. So, we proposed a two-stage flow inversion. First, we use the fixed-iteration technique to reduce the approximation error when estimating the flow velocity and get a basic inversion with the Euler method. This makes inversion approximates the authentic generation process. Then based on such inversion, we only need to add mild compensation at each denoising step to recover the exact original image. Since the compensation at each denoising step is small, the inversion trajectory is close to the model's original domain rather than overfitting one image, thereby preserving the editing ability.

In the investigation of the invariance control in MM-DiT, our key finding is that the replacement of the text features corresponding to the unedited prompt within the adaptive layer normalization (AdaLN) can simultaneously preserve the non-target contents while allowing rigid and non-rigid editing such as altering layout, quantity, and pose. This contrasts with the self-attention mechanism which tends to be indiscriminative to edited and unedited words in the text prompt and indistinctively injects the contents of the original image into the target, hindering some non-rigid editing types.
Consequently, our method can connect changes in the text prompt to the changes in AdaLN features and then guide image generation. Besides, the MM-DiT only has the self-attention that joins the text and image modalities together. This further makes it difficult to disentangle the target edited semantics from the invariance controlled by the attention mechanism and influence the desired editing (see Figure~\ref{fig:attn}). Thus, we choose the AdaLN as the main invariance control in MM-DiT, leaving the attention as a subtle complementary.

In summary, our contributions and findings include: 
\begin{itemize}
\item[$\bullet$] We unveil the deficiency and relation of Euler to DDIM inversion and propose a two-stage inversion to reduce the velocity approximation error and recover the exact image while enhancing the editing ability by making the inversion pivoting around the authentic generation process.

\item[$\bullet$] We introduce a flexible invariance control mechanism for the flow transformer (MM-DiT) based on adaptive layer normalization (AdaLN), which reconciles rigid and non-rigid editing and enables versatile editing types.
\item[$\bullet$] Experiments on various editing types validate that our method gives full play to the large prior of the flow transformer for versatile tuning-free image editing.

\end{itemize}

\section{Related Work}
\noindent\textbf{Image editing via diffusion models.}
Earlier editing methods fine-tune the model with the original image and the text prompt~\cite{kawar2023imagic,zhang2023forgedit}. Later approaches mainly fall into three categories. Truncation methods analyze latent noise predictions of the model and filter out undesired editing parts to preserve the non-target regions~\cite{brack2023sega,koo2024flexiedit}. The instruction-training methods~\cite{brooks2023instructpix2pix,zhang2024magicbrush,geng2024instructdiffusion,sheynin2024emu,meng2024instructgie,guo2024focus} collects a triplet of original and edited images and the editing instruction to re-train the diffusion model to be an editing model, which can be difficult to scale up. The inversion-based methods~\cite{xu2024inversion,hertz2022prompt,parmar2023zero,tumanyan2023plug,cao2023masactrl} are tuning-free and first convert the original image into the initial noisy latent and regenerate the edited image with the edited prompt while preserving the invariance (i.e., unchanged content from the original image). Others further improve the applicable ability by combining the large language model~\cite{huang2024smartedit,kwon2024zero} and enrich the editing forms~\cite{shi2024dragdiffusion,yang2024imagebrush,lin2024text,lu2024regiondrag}. ~\citet{hu2024latent} explores the FMs but based on U-ViT.~\cite{rout2024semantic,wang2024taming} explore the DiT yet they show limited tradeoffs in preserving invariance and editing in versatile non-rigid editing types.

\noindent\textbf{Inversion of diffusion for real image editing.}
The inversion that converts the image into the initial noisy latent is pivotal in tuning-free editing. One branch leverages DDIM~\cite{song2020denoising} to invert the ODE process. However, the classifier-free guidance (CFG) in editing deteriorates the accumulated error in reversing ODE. Various mitigation solutions explore optimizing the null-text embedding~\cite{mokady2023null}, adding corrections~\cite{ju2024pnp}, better estimation of latents~\cite{garibi2024renoise}, and negative prompt~\cite{miyake2023negative}. Another branch adopts the stochastic differential equation (SDE) instead of ODE to mitigate the error by manipulating random noise~\cite{brack2024ledits,huberman2024edit}. Besides these heuristic solutions, others explore the mathematical exact inversion by reformulating the sampling process~\cite{wallace2023edict,hong2024exact,wang2024belm}. Besides these nonparametric solutions, TurboEdit~\cite{wu2024turboedit} models the inversion process with a neural network to reduce the inversion time. Unlike these diffusion-oriented inversions, we specifically unveil the deficiency of flow inversion and study efficient inversion based on the velocity field of flow models.

\noindent\textbf{Invariance control in diffusion-based image editing}. The invariance control preserves the original image's non-target contents during editing. The instruction-based methods explicitly add the original image features as the condition and retrain the T2I model into a text-guided image-to-image (TI2I) model~\cite{brooks2023instructpix2pix,geng2024instructdiffusion,zhang2023adding,ye2023ip}. In the tuning-free paradigm, attention-based methods~\cite{hertz2022prompt,tumanyan2023plug,wang2024taming} inject attention maps from the original image to preserve the original contents but may struggle with non-rigid editing such as layout and pose. Later methods~\cite{cao2023masactrl,xu2023inversion,wang2024taming} inject K and V values of attentions in the later diffusion process to adapt to the layout change but this deteriorates the fidelity for large changes. The refinement approach~\cite{brack2023sega,koo2024flexiedit} filters out components of the predicted noise corresponding to the non-target regions. ~\citet{rout2024semantic} regulates the flow inversion with the original image. However, these approaches require careful tuning of hyperparameters for different editing types, and most are based on diffusion and U-Net. In contrast, we systematically investigate inversion for rectified flow and flexible invariance control based on MM-DiT to support rigid and non-rigid editing, enabling versatile editing types.

\begin{figure*}[ht!] 
\centering 
\includegraphics[width=0.99\linewidth]{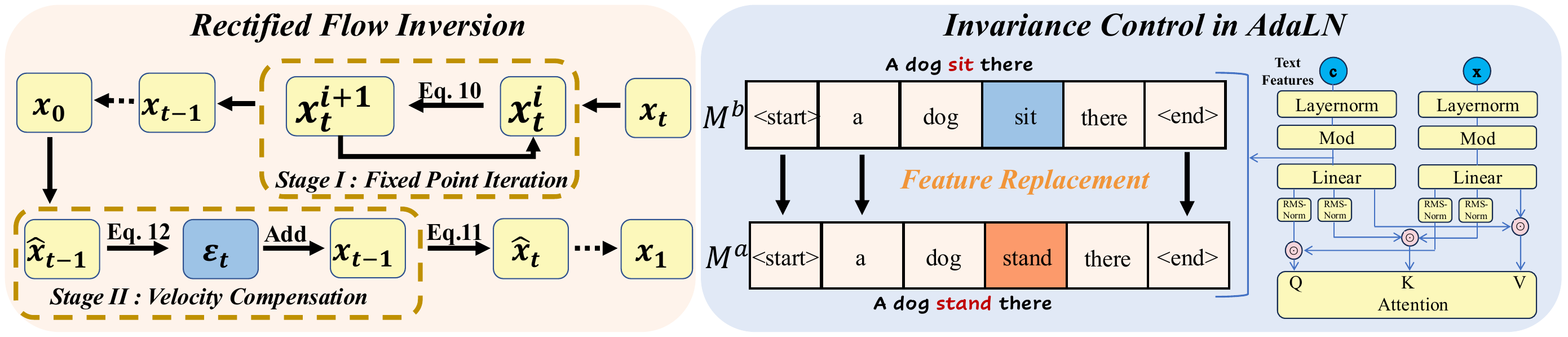}
\vspace{-3mm}
\caption{\textbf{Framework of rectified flow inversion and editing}. \textbf{Left}: The two-stage inversion for rectified flow. A basic inversion trajectory is first constructed to resemble the generation process, and then mild compensation is added to the velocity to recover the image exactly. \textbf{Right}: Invariance control with text feature replacement within AdaLN during sampling. In the target branch, the unchanged text features are replaced with features from the original image while the edited text features remain intact.} 
\label{frame}
\vspace{-3mm}
\end{figure*}

\section{Preliminary}
We aim to design a tuning-free framework based on the flow transformer for versatile editing. In Section 3, we briefly introduce the rectified flow and DDIM. Then, in Section 4, we analyze the deficiency of Euler inversion and propose the two-stage flow inversion. In Section 5, we present the image invariance investigation in MM-DiT centered on AdaLN. Last, we summarize the full algorithm.

\noindent\textbf{Rectified Flow.} Similar to the diffusion model~\cite{ho2020denoising,song2020score} that learns the probability paths between two distributions, rectified flow~\cite{liu2022flow,esser2024scaling} linearly interpolates the probability path between two observed distributions $\mathbf{x}_0 \sim p_0$ and $\mathbf{x}_1 \sim p_1$ in Eq.~\ref{eq:linear} and learns such straight probability transport paths with ODE in Eq.~\ref{eq:rf} where the transport is modeled by a time-aware velocity $v_{\theta}(\mathbf{x}_t, t)$. 
\begin{equation}
\label{eq:linear}
\mathbf{x}_t=t\mathbf{x}_1+(1-t)\mathbf{x}_0, \quad t\in[0,1].
\end{equation}
\begin{equation}
\label{eq:rf}
\mathrm{d}\mathbf{x}_t=v_{\theta}(\mathbf{x}_t, t)\mathrm{d}t,\quad \mathbf{x}_0\thicksim p_0,\quad t\in[0,1].
\end{equation}
In practice, the velocity $v_{\theta}$ of rectified flow is parameterized by the weights $\theta$ of a neural network and can be straightforwardly derived from differentiating Eq.~\ref{eq:linear} as $\mathbf{x}_1 - \mathbf{x}_0$. The training objective is to directly regress the velocity field in with the least squares loss Eq.~\ref{eq:rf_train} and $\mathbf{x}_t$ in Eq.~\ref{eq:linear}.
\begin{equation}
\label{eq:rf_train}
\mathcal{L} = \mathbb{E}_{t \sim \mathcal{U}[0,1], \mathbf{x}_1 \sim p_1}\left[\left\|(\mathbf{x}_1-\mathbf{x}_0)-v_{\theta}(\mathbf{x}_t,t)\right\|^2\right].
\end{equation}
Generally, $p_0$ is chosen as the standard Gaussian $\mathcal{N}(0, I)$, and $p_1$ is the target data distribution. The sampling process is to start as $\mathbf{x}_0$ and integrate the ODE from $t: 0 \to 1$ to yield a data sample $\mathbf{x}_1$, which is practically implemented by solving the discrete ODE by Euler method in Eq.~\ref{eq:euler} where $\sigma_t$ is the discrete timestep.
\begin{equation}
\label{eq:euler}
\mathbf{x}_{t+1} = \mathbf{x}_{t} + (\sigma_{t+1} - \sigma_{t}) v_{\theta}(\mathbf{x}_t,t)
\end{equation}

\noindent\textbf{DDIM.} DDIM~\cite{song2020denoising} is the implicit probabilistic model trained with DDPM~\cite{ho2020denoising} objective and has the formulation below where $\mathbf{x}_t$ is also the noisy data sample, $\mathbf{s}_\theta(\mathbf{x}_t,t)$ is the noise prediction, $\mathbf{f}_\theta( \mathbf{x}_{t}, t)$ is the predicted original data sample.
\begin{equation}
\label{eq:ddim1}
\mathbf{x}_{t+1} = \sqrt{\alpha_{t+1}} \mathbf{f}_\theta( \mathbf{x}_{t}, t) + \sqrt{1-\alpha_{t+1}} \mathbf{s}_\theta(\mathbf{x}_t,t)
\end{equation}
\begin{equation}
\label{eq:ddim2}
\mathbf{f}_\theta( \mathbf{x}_{t}, t) = \frac{\mathbf{x}_{t} - \sqrt{1-\alpha_{t}} \mathbf{s}_\theta(\mathbf{x}_t,t)} {\sqrt{\alpha_t}}
\end{equation}

\section{Inversion for Rectified Flow}
The inversion of the sampling process from the data sample $\mathbf{x}_1$ back to $\mathbf{x}_0$ can be solved by reverse ODE. In rectified flow, the \textbf{exact} Euler inversion is Eq.~\ref{eq:eu_inv}. However, since we move backward, we only have $\mathbf{x}_{t+1}$ but not $\mathbf{x}_t$. So, we approximate $\mathbf{x}_{t}$ with $\mathbf{x}_{t+1}$ in Eq.~\ref{eq:eu_inv2}. This introduces the \textbf{approximation error} at every step and accumulates to deviate the initial latent $\mathbf{x}_{0}$. The same applies to DDIM inversion.
\begin{align}
\label{eq:eu_inv}
\mathbf{x}_{t} &= \mathbf{x}_{t+1} + (\sigma_{t} - \sigma_{t+1}) v_{\theta}(\mathbf{x}_t,t) \\
\label{eq:eu_inv2}
&\approx \mathbf{x}_{t+1} + (\sigma_{t} - \sigma_{t+1}) v_{\theta}(\mathbf{x}_{t+1},t)
\end{align}

\subsection{Deficiency and Relation of Euler to DDIM} 
\label{inv:con}
\begin{figure}[H] 
\centering 
\includegraphics[width=0.98\linewidth]{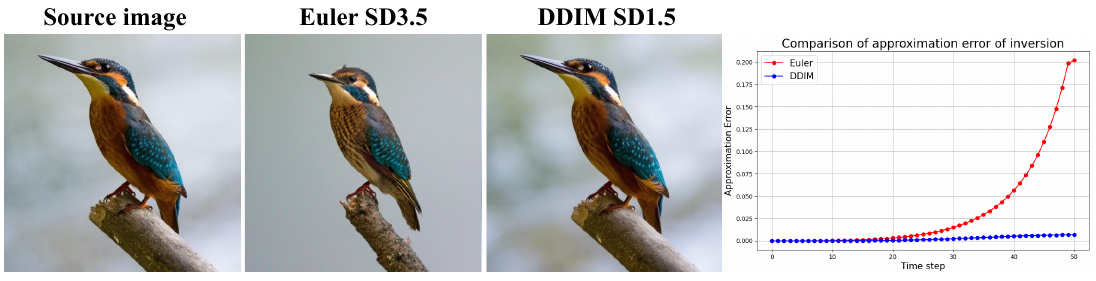}
\vspace{-3mm}
\caption{\textbf{Comparison of Euler inversion for rectified flow and DDIM inversion.} With the same inversion steps as 50, the Euler is evaluated on SD3.5 while the DDIM is evaluated on SD1.5.}
\label{com:inv}
\vspace{-2mm}
\end{figure}

Essentially, if we rescale $\mathbf{x}_t$ with $\frac{\mathbf{x}_t}{\sqrt{\alpha_t}}$ and rewrite Eq.~\ref{eq:ddim1}, DDIM can also be treated as the first-order ODE solver as Euler above. The corresponding inversion is Eq.~\ref{eq:ddim_ode}.
\begin{align}
\label{eq:ddim_ode}
\frac{\mathbf{x}_{t}}{\sqrt{\alpha_{t}}} &\approx \frac{\mathbf{x}_{t+1}}{\sqrt{\alpha_{t+1}}} + (\sqrt{\frac{1-\alpha_t}{\alpha_t}} - \sqrt{\frac{1-\alpha_{t+1}}{\alpha_{t+1}}}) \mathbf{s}_\theta(\mathbf{x}_{t+1},t)
\end{align}
However, despite the similar formulation of Euler and DDIM inversion, the performance of Euler inversion is significantly worse than DDIM inversion, even if the Euler inversion is evaluated on the more powerful Stable Diffusion 3.5 vs SD 1.5 as shown in Figure~\ref{com:inv}. Such phenomena are not expected and also make optimization methods~\cite{mokady2023null,ju2024pnp,brack2024ledits} in diffusion invalid for flow inversion.

We argue this is mainly due to that the Euler inversion for the rectified flow transformer suffers more from the approximation error. To reduce it, we need to mitigate the gap between $v_{\theta}(\mathbf{x}_{t+1},t)$ and $v_{\theta}(\mathbf{x}_{t},t)$. Idealy, we expect to use the exact inversion in Eq.~\ref{eq:eu_inv} to get the initial noisy latent $\mathbf{x}_0$ and the transition trajectory $\{\mathbf{x}_1,...,\mathbf{x}_t\}$. Such an inversion trajectory is close to the trajectory in the generating process and requires minimal optimization. Thus, it fits the prior distribution of the model and has good editing ability.

\subsection{Two-stage Flow Inversion}
Our principle is to find an initial latent and inversion trajectory that is as close to the authentic generation process as possible, trying to avoid only overfitting the original image. This has two-fold benefits: 1. \textbf{Editing friendly}. The latent and trajectory fit into the prior distribution of the model and text-to-image alignment is authentic so that the image can be easily manipulated by changing the text prompt. 2. \textbf{Easy to preserve invariance and fidelity}. If the inversion trajectory deviates from the model prior, it requires more effort, such as self-attention injection to preserve the invariance and fidelity. However, injecting too much self-attention from the original image can hinder the editing. Thus, we propose a two-stage flow inversion. First, we construct a pivotal inversion close to the generating process. Second, we manually add compensation at every timestep to eliminate the left mild error.

\begin{figure}[htb]
\centering 
\includegraphics[width=0.95\linewidth]{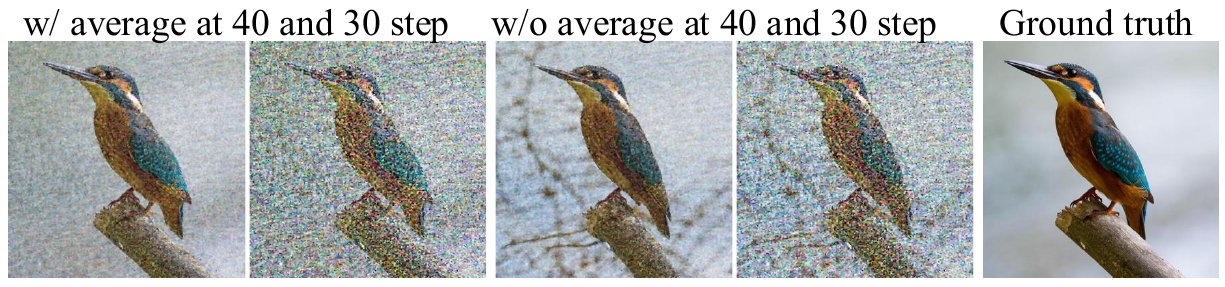}
\vspace{-3mm}
\caption{\textbf{Visualization of latents $\mathbf{x}_t$ w / w.o averaging}. } 
\label{fig:average}
\vspace{-3mm}
\end{figure}

\noindent\textbf{Stage I: Fixed point iteration with stable velocity.} We aim to use the Eq.~\ref{eq:eu_inv} to get the exact $\mathbf{x}_{t}$ without the approximation. To achieve this, we need to improve the estimation of $v_\theta(\mathbf{x}_t, t)$ instead of using the approximated $v_\theta(\mathbf{x}_{t+1}, t)$. Note that the input to $v_\theta(\mathbf{x}_t, t)$ and output of Eq.~\ref{eq:eu_inv} are both $\mathbf{x}_t$. Inspired by the fixed-point technique~\cite{garibi2024renoise,granas2003fixed}, we can iteratively apply Eq.~\ref{eq:eu_inv} to $\mathbf{x}_t$ itself and better estimate the $v_\theta(\mathbf{x}_t, t)$. Concretely, we first initialize $\mathbf{x}_{t}^1$ with $\mathbf{x}_{t+1}$ and iteratively apply the following equation to get a series estimation of $\{\mathbf{x}_t^i\}_{i=1}^{I}$. 
\begin{equation}
\label{eq:fix}
\mathbf{x}_{t}^{i+1} = \mathbf{x}_{t+1} + (\sigma_{t} - \sigma_{t+1}) v_{\theta}(\mathbf{x}_t^{i},t) 
\end{equation}
Besides, ideally, the velocity $v_{\theta}$ is $\mathbf{x}_1 - \mathbf{x}_0$ according to Eq.~\ref{eq:rf_train} that is a stable value. This also strengthens the motivation of fixed-point iteration, and we can use this property to average the velocities to get a more stable velocity prediction. Practically, this is also equivalent to averaging the latent series $\{\mathbf{x}_t^i\}_{i=1}^{I}$. We compare the latents with and without averaging at different steps in Figure~\ref{fig:average} and show that averaged latents are closer to the original image since the fixed-point iteration may not converge monotonically and wander around the true fixed point.
\vspace{-3mm}

\noindent\textbf{Stage II: Velocity compensation in editing.} In Stage I, we get an inversion trajectory $\{\mathbf{x}_t\}_{t=1}^{T}$ that significantly reduces the approximation error and is close to the generation process. However, since the fixed-point iteration is a numerical method, the error may still exist when using the inverted $\mathbf{x}_0$ to recover $\mathbf{x}_1$. To exactly recover the original image $\mathbf{x}_1$ with the inversion trajectory $\{\mathbf{x}_t\}_{t=1}^{T}$, at every timestep, we manually calculate and add compensation $\epsilon_t$ for the velocity during the generating process (editing process) so that the recovery exactly follow the inversion trajectory.
\begin{equation}
\label{eq:compen1}
\mathbf{\hat{x}}_{t+1} = \mathbf{x}_{t} + (\sigma_{t+1} - \sigma_{t}) v_{\theta}(\mathbf{x}_{t},t)
\end{equation}
\begin{equation}
\label{eq:compen2}
\epsilon_t = \mathbf{x}_{t+1} - \mathbf{\hat{x}}_{t+1}
\end{equation}
Note that this stage is done during editing but not traversing the denoising process again. The algorithm is summarized in Algorithm~\ref{algo:optimization}. In practice, the inversion pivots around the authentic generation process with mild deviation, which recovers the exact image and provides good editing ability.

\begin{algorithm}[t]
\small
    \caption{Two-stage Flow Inversion and Editing}
    \label{algo:optimization}
    \begin{algorithmic}[1]
    \STATE \textbf{Input}: origin image $\mathbf{x}_1$, iteration steps $I$, inversion steps $T$, velocity model $v_\theta$, before after prompts $\mathcal{P}_b$, $\mathcal{P}_a$, timestep $S$.
    \STATE \texttt{Stage I: Fixed-point Inversion}
    \FOR{$t=1,\cdots,0$ ($T$ steps)}
        \STATE $\mathbf{x}_{t-1}^0 = \mathbf{x}_{t}$
        \FOR{$i=1,\cdots,I$}
        \STATE $\mathbf{x}_{t-1}^{i} = \mathbf{x}_{t} + (\sigma_{t-1} - \sigma_{t}) v_{\theta}(\mathbf{x}_{t-1}^{i-1},t-1, \mathcal{P}_s)$
        \ENDFOR
        \STATE $\mathbf{x}_{t-1} = \frac{1}{I}\sum_1^I \mathbf{x}_{t-1}^i$   \hfill $\triangleright$ \textit{Average velocities}
    \ENDFOR
    \STATE Get the inversion trajectory $\{\mathbf{x}_t\}_{t=1}^{T}$
    \STATE \texttt{Stage II: Velocity Compensation}
    \STATE Initialize before $\mathbf{x}^b_{0}$ and after $\mathbf{x}^a_{0}$ with $\mathbf{x}_{0}$.
    \FOR{$t=0,\cdots,1$ ($T$ steps)}
        \STATE \textbf{if} $t < S$ \textbf{then} $\mathbf{Map}(M^b, M^a, \mathcal{P}_b, \mathcal{P}_a)$
        \STATE $\mathbf{\hat{x}}^b_{t+1} = \mathbf{x}^b_{t} + (\sigma_{t+1} - \sigma_{t}) v_{\theta}(\mathbf{x}^b_{t},t, \mathcal{P}_b)$
        \STATE $\epsilon_t = \mathbf{x}^b_{t+1} - \mathbf{\hat{x}}^b_{t+1}$
        \STATE $\mathbf{x}^a_{t+1} = \mathbf{x}^a_{t} + (\sigma_{t+1} - \sigma_{t}) v_{\theta}(\mathbf{x}^a_{t},t, \mathcal{P}_a) + \epsilon_t$
    \ENDFOR
    
\STATE \textbf{Output}: The edited after image $\mathbf{x}^a_1$
\end{algorithmic}
\end{algorithm}

\begin{figure*}[htb] 
\centering 
\includegraphics[width=0.95\textwidth]{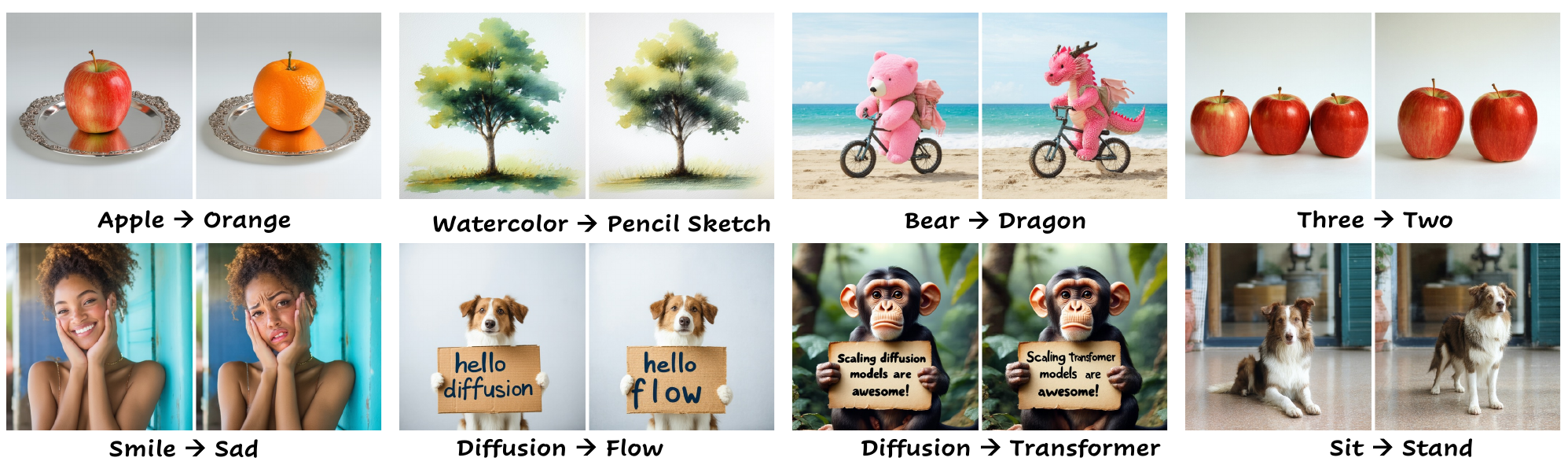}
\vspace{-3mm}
\caption{\textbf{Invariance control of AdaLN in different editing scenarios}. We show the proposed module can adapt to versatile editing types with high fidelity, including \textit{non-rigid editing} such as altering quantity, facial expression, pose, visual text, etc.} 
\label{fig:adaln}
\vspace{-3mm}
\end{figure*}

\section{Flexible Invariance Control with AdaLN}
\noindent\textbf{Motivation.} The invariance control aims to preserve the non-target contents in the original image while allow the text prompt to flexibly edit target regions. Previous methods based on the U-Net mainly utilize the attention map injection~\cite{cao2023masactrl,hertz2022prompt,xu2023inversion} and filtering non-target contents in predicted noise or frequency domain~\cite{koo2024flexiedit,brack2023sega} to adapt to various editing types. However, these still struggle to reconcile diverse editing types, especially for rigid and non-rigid editing. The reason is that self-attention injections such as KV injection and frequency filtering do not \textit{directly} correspond to the text and are not well aligned with text changes. Thus, such invariance preservation may counteract some text-guided changes such as layout, quantity, and large global changes. Another corresponding issue is that these mechanisms require tuning hyperparameters for different editing types, which can be sensitive and inefficient.

Considering that the flow transformer has a broader and more diverse text-to-image generation ability and MM-DiT joins the text and image features in self-attention, the new architecture, and the improved generation ability motivate us to design a unified invariance control mechanism for both rigid and non-rigid editing that better aligned with the text changes and enable diverse editing types without intricate tuning for each type.

\noindent\textbf{Module design.} In the investigation of MM-DiT, our key observation is that the text features within the adaptation layer normalization (AdaLN) closely correspond to the image semantics such as pose, quantity, and object types. Thus, we can inject unchanged text features from the original text into the target and keep the features of the edited text intact. This will only change regions indicated by text changes. Concretely, as depicted in Figure~\ref{frame}, we denote the before and after text features of the AdaLN before the attention in the MM-DiT block as $M^b, M^a \in \mathbb{R}^{j \times d}$ where $j$ represent the number of tokens of the text prompt and $d$ represent the text feature dimension. We define the token-aware Map function $\mathbf{Map}(M^b, M^a, \mathcal{P}_b, \mathcal{P}_a)$ to replace the features of unedited tokens in the after prompt $\mathcal{P}_a$ with corresponding tokens in the before prompt $\mathcal{P}_b$. Such operation is conducted starting from the beginning to the timestep $S$.
\begin{equation}
\label{eq:map}
\mathbf{Map}(M^b, M^a, \mathcal{P}_b, \mathcal{P}_a) := \left\{
\begin{aligned}
\hat{M}^a & , if\ t < S, \\
M^a & , otherwise.
\end{aligned}
\right.
\end{equation}

\begin{figure}[htb]
\centering 
\includegraphics[width=0.98\linewidth]{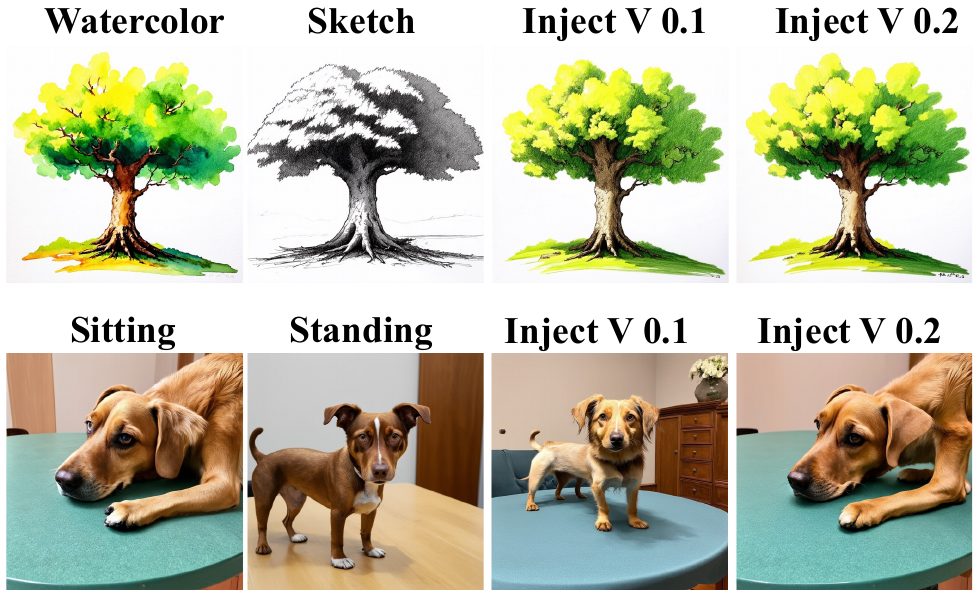}
\vspace{-3mm}
\caption{\textbf{Invariance control of self-attention with different time periods}. We inject different components (Q,K,V) of self-attention with different time periods to verify the ability to preserve the unedited contents (Appendix Sec:~\ref{app:self-attn}). Here we show injecting V value more than 20\% time steps can hinder the non-rigid editing `\texttt{sitting to standing}'.} 
\label{fig:attn}
\vspace{-3mm}
\end{figure}

The proposed module connects the features to the changes in the text prompt and can flexibly control the invariance without hindering the editing. In practice, we build such an editing module based on the MM-DiT Stable Diffusion 3.5. We show the potential of editing in various scenarios in Figure~\ref{fig:adaln}. The results show that the replaced text features $\hat{M}^a$ in AdaLN adapt to different editing scenarios.
To complete our argument, we also present the investigation of Self-Attention of MM-DiT for invariance control in Figure~\ref{fig:attn}. The observations are similar to previous experiences in U-Net. Attention injection in too many timesteps may prevent some non-rigid editing effects. In contrast, we keep the injection timestep $S$ the same for all editing types, and such a unified and stable timestep does not hinder non-rigid editing. Please refer to Appendix Sec:~\ref{app:self-attn} for details.

\section{Experiments}
We evaluate our methods in various editing types and scenarios and further analyze the effects of two-stage inversion and iteration steps with qualitative and quantitative results. Please also see the Appendix for more results.

\begin{figure*}[htb] 
\centering 
\includegraphics[width=0.9\linewidth]{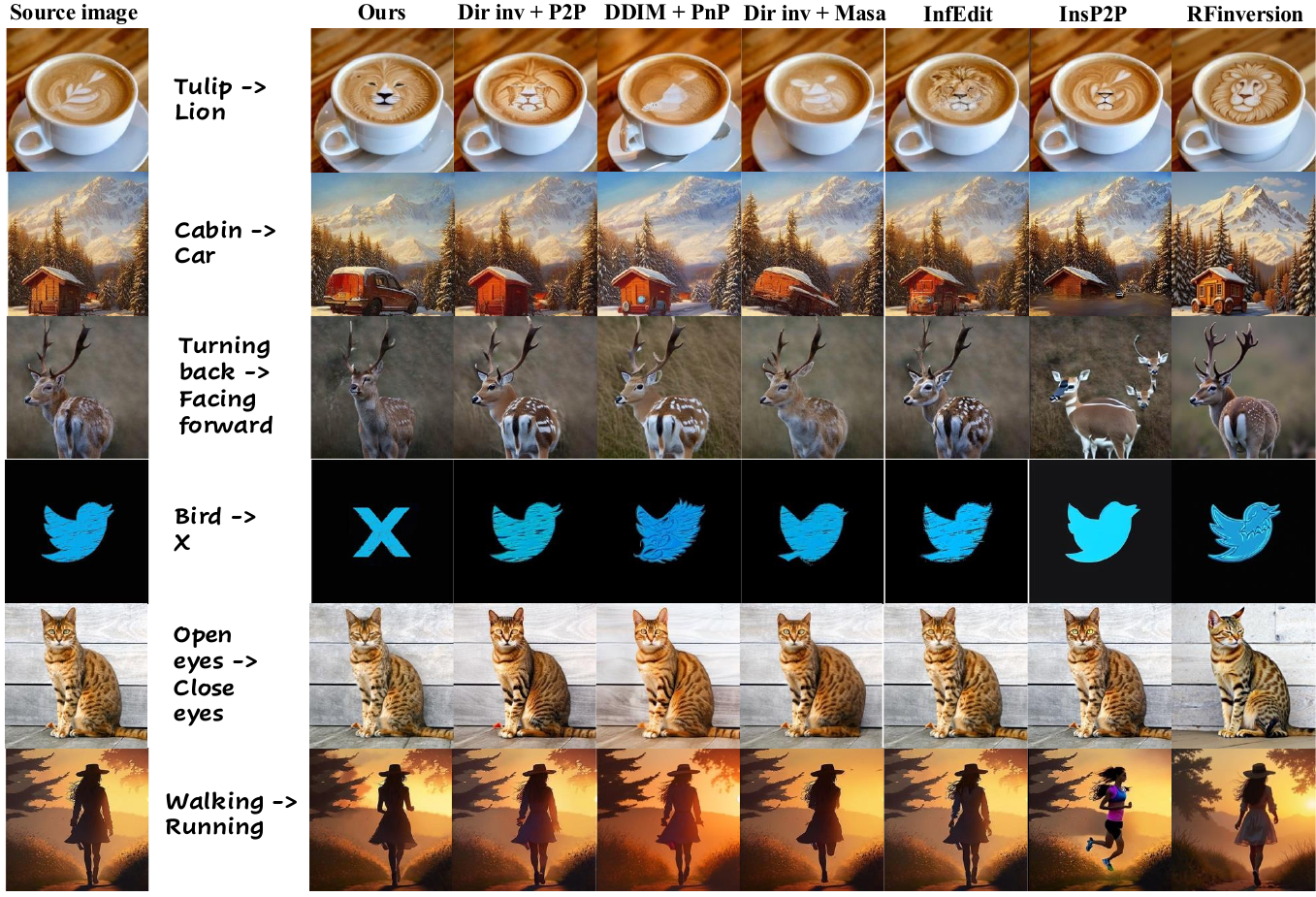}
\vspace{-3mm}
\caption{\textbf{Qualitative comparisons on real image editing}. With the proposed inversion and invariance control, our method can edit real images in different scenarios with different editing types, achieving high fidelity for different levels of geometric changes.} 
\label{fig:real_comp}
\vspace{-3mm}
\end{figure*}

\subsection{Setup}
\noindent\textbf{Baselines and implementations.} We mainly compare our methods with previous state-of-the-art turning-free image editing methods: P2P~\cite{hertz2022prompt}, PnP~\cite{tumanyan2023plug}, MassCtrl~\cite{cao2023masactrl}, InfEdit~\cite{xu2023inversion}, RFinv~\cite{rout2024semantic}. The first four use attention maps in U-Net to control the invariance for editing. Specifically, MasaCtrl and InfEdit address the issue of non-rigid editing. RFinv~\cite{rout2024semantic} is close to our work exploring flow-based models for tuning-free editing, but is based on Flux-dev. Additionally, we also include the instruction-based editing method InsP2P~\cite{brooks2023instructpix2pix}, which is retrained with the triplet of before-and-after image pairs and the editing instruction. We follow their official implementations for evaluation for all methods. We implement our method based on the Stable Diffusion 3.5 and use the basic Euler sampler of rectified flow for sampling and inversion. We use 30 steps for inversion and choose the CFG as 1 and 2 for inversion and editing respectively. We set the fixed-point iteration number as 3. For methods based on diffusion U-Net and DDIM sampling, we use 50 as inversion steps and choose CFG as 7, which follows their default settings.

\noindent\textbf{Evaluation datasets and metrics.} We evaluate our method based on the PIE benchmark~\citep{ju2024pnp} which includes 700 natural and artificial images in total. The dataset covers a wide range of editing types and provides detailed metrics to evaluate both the editing ability and invariance preservation ability at the same time. Concretely, for invariance preservation, the metrics include PSNR, LPIPS, MSE, and SSIM. Note that the numbers of these metrics reported in this paper are scaled. For the editing ability, the metric calculates the CLIP similarity between the whole image and the target prompt (Whole) and between the target regions and the target text prompt (Edited).

\subsection{Comparison with previous editing methods}
\noindent\textbf{Real image editing and fidelity.} We compare our method with others in real image editing. As shown in Figure~\ref{fig:real_comp}, we evaluate methods with different editing types that represent different levels of geometric changes. Our method can better preserve non-target content and achieve high fidelity. This validates the generalization ability of our method in real image editing. In contrast, methods that use attention injection to preserve the invariance cannot reconcile different editing types. Specifically, P2P and PnP utilize the cross and self-attention injection to preserve better structure information but they also sacrifice the ability to edit the layout and pose of objects. MasaCtrl and InfEdit inject the KV in the later diffusion steps to allow editing of the layout and pose since the structure information tends to form in the early steps. However, their fidelity is affected when facing large changes. For example, MasaCtrl is limited in object replacement. The instruction-based InsP2P is flexible to all editing types but inferior in preserving the structure information since the training data are difficult to construct and may not be accurate, which influences editing fidelity. Please also see the Appendix for more qualitative results.

\begin{table}[H]
\small
\centering
\tabcolsep=0.1cm
\caption{\textbf{Quantitative comparisons in PIE benchmark.}}
\vspace{-3mm}
\resizebox{1\linewidth}{!}{
\begin{tabular}{@{} l S S S S S S S @{}}
\toprule
& \multicolumn{1}{c}{Structure} & \multicolumn{4}{c}{Background Preservation} & \multicolumn{2}{c}{CLIP Similarity} \\
\cmidrule(lr){2-2} \cmidrule(lr){3-6} \cmidrule(lr){7-8}
\scriptsize \textbf{Method} & \scriptsize{$\textbf{Distance}$ $\downarrow$} & \scriptsize{$\textbf{PSNR}$ $\uparrow$} & \scriptsize{$\textbf{LPIPS}$ $\downarrow$} & \scriptsize{$\textbf{MSE}$ $\downarrow$} & \scriptsize{$\textbf{SSIM}$ $\uparrow$} & \scriptsize{$\textbf{Whole}$ $\uparrow$} & \scriptsize{$\textbf{Edited}$ $\uparrow$} \\
\midrule
PnP  & 24.29 & 22.46 & 106.06 & 80.45 & 79.68 &\underline{25.41} &\textbf{22.62} \\
P2P  &\textbf{13.44} &\textbf{27.03} &\textbf{60.67} &\textbf{35.86} &\underline{84.11} &24.75 &21.86 \\
MasaCtrl  & 24.70 & 22.64 & 88.79 &  81.09 & 80.76 & 24.38 & 21.35 \\
InsP2P  &57.91 &20.82 &158.63 &227.78 &76.26 &23.61 &21.64 \\
InfEdit  & 24.70 & 26.31 & 87.94 &  75.19 & 81.33 & 23.67 & 21.86 \\
RFinv  &32.62  &22.03  &159.62  &96.01  &73.26  &24.89  &21.89 \\
Ours & \underline{18.17} & \underline{26.62} & \underline{80.55} & \underline{40.24} & \textbf{91.50} & \textbf{25.74} & \underline{22.27} \\
\bottomrule
\end{tabular}
}
\label{tab:main}
\end{table}

\vspace{-2mm}
\noindent\textbf{Quantitative comparison.} The quantitative results are summarized in Table~\ref{tab:main}. The comparison methods are implemented with the most suitable diffusion-based inversions for high performance. P2P and MasaCtrl are based on Direct Inversion (Dir Inv)~\cite{ju2024pnp}. Our method achieves good performance in most metrics. Concretely, the results show that P2P has a better ability to preserve the contents of the source image but this also hinders the ability to edit images. As shown in CLIP scores, the P2P CLIP scores are relatively lower. In contrast, the PnP has a higher CLIP Edited score but the other metrics such as PSNR and LPIPS are worse. This indicates that PnP cannot preserve the invariance information well. In summary, our method achieves good results in both background preservation and editing. This means that our method can make accurate editing without sacrificing either the invariance or the editing ability.

\noindent\textbf{Non-rigid editing.} Furthermore, to validate that our method can reconcile rigid and non-rigid editing scenarios, we specifically evaluate methods in non-rigid editing such as shape and pose with numerical experiments. We summarize the detailed results in the Appendix, and the comparison shows that our method outperforms others. This supports the motivation of our proposed invariance control module. The invariance control based on the AdaLN is aligned with the text changes so that the edited image can better adapt to non-rigid changes.

\begin{figure}[t]
\centering 
\includegraphics[width=0.99\linewidth]{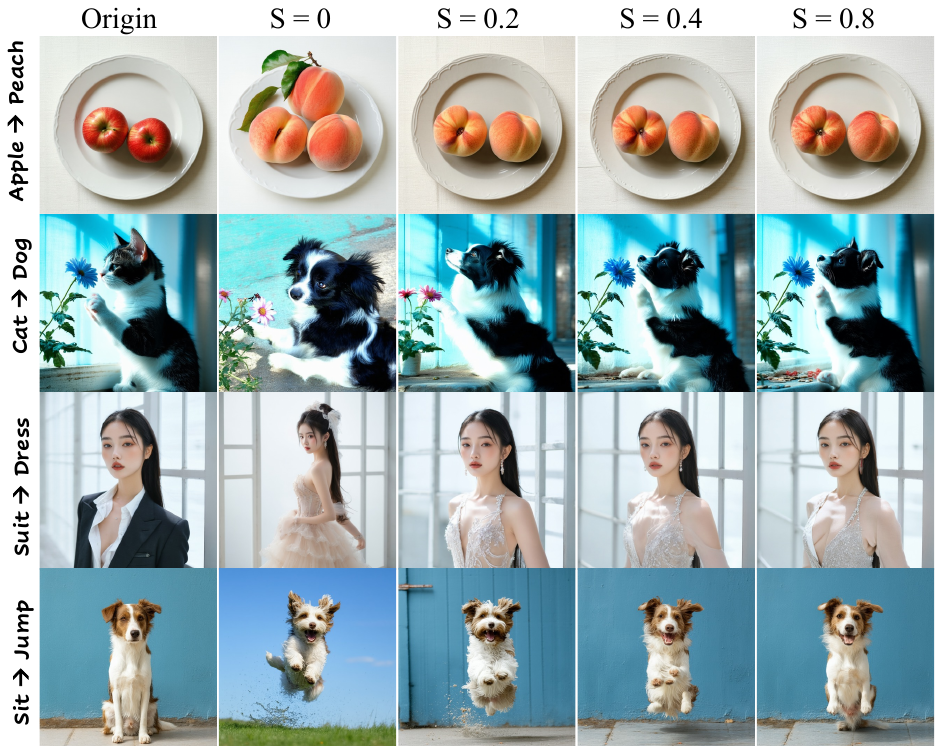}
\vspace{-3mm}
\caption{\textbf{Ablation study of invariance control based on AdaLN under different periods}. Injection with a large period $S$ still reflects the rigid and non-rigid edited effect.} 
\label{fig:ada_time}
\vspace{-4.5mm}
\end{figure}

\subsection{Ablation Study and Analysis}
We ablate the proposed inversion and invariance control module to validate their effectiveness. We randomly sampled 150 images from the PIE benchmark for ablation study.

\begin{figure}[t] 
\centering 
\includegraphics[width=0.98\linewidth]{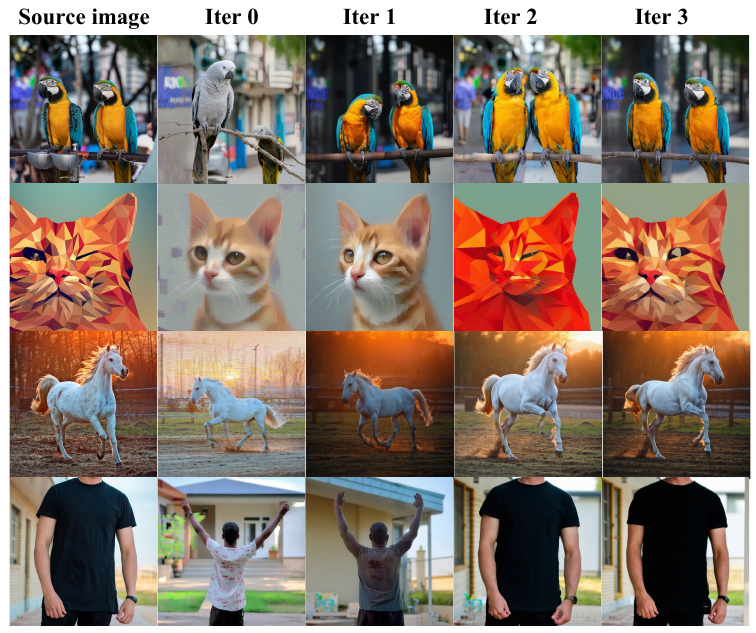}
\vspace{-3mm}
\caption{\textbf{Qualitative results of flow inversion under different iterations}. Fixed-point iteration can make the recovered image approximate the original image and then the left mismatch can be eliminated by the velocity compensation at each timestep.} 
\label{fig:inv_inter}
\vspace{-4mm}
\end{figure}

\noindent\textbf{Number of fixed-point iterations.} We evaluate the effectiveness of the fixed-point iteration with different iteration numbers. The results are summarized in Table~\ref{tab:abla}. Iter 0 means using the plain Euler inversion. The results show that the fixed-point iteration can improve the performance of preserving the unedited contents since the inversion becomes more accurate. Besides, a larger iteration number can improve the performance, which is expected since more iterations improve the accuracy of inversion. However, we also observe that the benefit of using large numbers on invariance preservation is marginal but this helps the editing since the CLIP scores are increased. We think the large number makes the inversion trajectory better converge to the authentic generation process and benefits editing.
We also show the qualitative results of inversion under different iteration numbers in Figure~\ref{fig:inv_inter}. The results show that a large iteration number can facilitate the flow inversion. However, the larger iteration number also increases the computation load which needs to be considered for practical usage.

\begin{table}[t]

\small
\centering
\tabcolsep=0.1cm
\caption{\textbf{Ablation study on the fixed point iteration number and velocity compensation.} We choose different iteration numbers to validate their effectiveness in editing.}
\vspace{-3mm}
\label{tab:abla}
\resizebox{1\linewidth}{!}{
\begin{tabular}{@{} l S S S S S S S @{}}
\toprule
& \multicolumn{1}{c}{Structure} & \multicolumn{4}{c}{Background Preservation} & \multicolumn{2}{c}{CLIP Similarity} \\
\cmidrule(lr){2-2} \cmidrule(lr){3-6} \cmidrule(lr){7-8}
\scriptsize \textbf{Method} & \scriptsize{$\textbf{Distance}$ $\downarrow$} & \scriptsize{$\textbf{PSNR}$ $\uparrow$} & \scriptsize{$\textbf{LPIPS}$ $\downarrow$} & \scriptsize{$\textbf{MSE}$ $\downarrow$} & \scriptsize{$\textbf{SSIM}$ $\uparrow$} & \scriptsize{$\textbf{Whole}$ $\uparrow$} & \scriptsize{$\textbf{Edited}$ $\uparrow$} \\
\midrule
Iter 0   &48.49  &20.78 &199.93 &102.82 &80.59 &24.14 &21.39    \\ 
Iter 1   &15.29  &26.98 &67.80 &35.68 &93.89 &24.31 &21.47    \\ 
Iter 2   &15.02  &27.28 &66.85 &34.98 &93.79 &24.30 &21.57       \\ 
Iter 3   &14.92  &27.51 &66.48 &34.06  &94.40 &24.48 &21.73       \\
\bottomrule
\end{tabular}%
}
\vspace{-2mm}
\label{tab:abla}

\end{table}

\noindent\textbf{Velocity compensation.} The velocity compensation eliminates the leftover error at each timestep when recovering the original image. The qualitative results in Figure~\ref{fig:inv_inter} show that the left error is not large and the velocity compensation can recover the exact original image and improve the fidelity of editing. For the detailed qualitative results and analysis, please refer to Appendix.

\noindent\textbf{Replacement of AdaLN within different time periods.} To investigate the relation between invariance preservation with different time periods. We show the editing results with AdaLN invariance control within different time periods. The results in Figure~\ref{fig:ada_time} show that after a certain timestep, the edited image does not change a lot even if we increase the time period. This indicates that the hyperparameter replacement timestep $S$ is not sensitive. Besides, the results also show that applying text feature replacement within the AdaLN does not hinder non-rigid editing, which is better than the attention mechanism. So, we can set $S$ to a relatively stable number in practice.

\noindent\textbf{Limitation.} Tuning-free image editing is a system consisting of inversion and invariance control. One limitation is that the invariance control based on AdaLN depends on the authentic inversion resembling the generation process. If the real image is out of the model's domain and the inversion deviates too much from the generation process, the module cannot precisely control the contents since the inversion does not fit the model's prior distribution and text-to-image alignment is mismatched. Another limitation is the computational efficiency of the fixed-point iteration since this requires inferencing the transformer more times but in practice, the overhead is not large. 
We present the failure cases in the Appendix. The out-of-domain images with poor inversion cannot be well preserved and manipulated as expected. The editing results are determined by the model's prior, which is different from the original image.

\section{Conclusion}
We investigated tuning-free image editing based on the flow
transformer. We provide systematic analysis for the inversion of rectified flow and invariance control based on the MM-DiT. We propose a two-stage inversion strategy to reduce the approximation error of estimating the velocity field and try to find an inversion approximating the authentic generation process to enhance the editing ability. Besides, we propose the invariance control based on the AdaLN to reconcile rigid and non-rigid editing. The AdaLN control mechanism discriminately preserves the original image features according to the text changes, enabling more diverse editing types, especially for rigid and non-rigid editing types. Both qualitative and quantitative experiments validate the proposed framework can adapt to versatile editing types and fully leverage the large generation prior of the flow transformer for tuning-free image editing. We hope our work can inspire thinking over flexible and accurate invariance control in large-scale T2I models to take full advantage of the generation prior for downstream tasks.

{
    \small
    \bibliographystyle{ieeenat_fullname}
    \bibliography{main}
}


\clearpage
\clearpage
\setcounter{page}{1}
\maketitlesupplementary

The supplementary is organized as follows: 
\begin{itemize}
    \item Additional detailed related work: 1. text-to-image flow-based models. 2. invariance control in diffusion-based image editing.
    \item Analysis of approximation error in Euler inversion.
    \item Versatile image editing with AdaLN invariance control.
    \item Investigation of attention-based invariance control in flow transformer.
    \item Additional non-rigid editing results.
    \item Failure case study.
\end{itemize}

\section{Additional related work}
We present the detailed discussion as the extension for the related work in the main text.

\noindent\textbf{Text-to-Image Flow-based models.}
Flow-based models~\cite{liu2022flow,albergo2022building,lipman2022flow} interpolate the probability transition path between two data distributions via the ordinary differential equation (ODE) and learn the conditional velocity field of the transition path. Such a formulation also implies the diffusion models that use Gaussian probability paths. The advantage over diffusion models is to allow faster simulation of the probability flow ODEs, which induces fewer sampling steps. Later works further rectify and optimize the non-optimal transition paths~\cite{pooladian2023multisample,ma2024sit,hu2024latent,xie2024reflected}. Driven by these theoretical benefits, recent large-scale text-to-image models such as Stable Diffusion 3~\cite{esser2024scaling} and Flux implement flow matching with the diffusion transformer architecture (DiT)~\cite{peebles2022scalable} and achieve new state-of-the-art text-to-image synthesis. However, recent image editing (especially tuning-free editing) approaches are mainly based on diffusion and U-Net~\cite{ronneberger2015u} whereas flow matching and transformer lack exploration. This paper investigates the flow-transformer models as the foundation for tuning-free image editing. We analyze flow inversion and image invariance control based on transformer architecture in editing.

\noindent\textbf{Invariance control in diffusion-based image editing}. The invariance control preserves the original image's unedited contents in diffusion-based image editing. The instruction-based methods explicitly add the original image features as the condition and retrain the T2I model into a text-guided image-to-image (TI2I) model. For example, the InsP2P~\cite{brooks2023instructpix2pix} and InsDiffusion~\cite{geng2024instructdiffusion} concate the latent of the original image with the noisy latent. The IP-adapter~\cite{ye2023ip} and T2I-adapter~\cite{mou2024t2i} add the extra branch to the U-Net to inject features into the U-Net of diffusion. In the tuning-free paradigm, P2P found injecting the cross-attention corresponding to the text prompt can main the unedited contents. Furthermore, P2P-zero~\cite{hertz2022prompt} and Plug-and-Play~\cite{tumanyan2023plug} explore the self-attention to preserve the invariance during editing. However, these attention-based methods struggle with non-rigid editing such as changing the layout. MasaCtrl~\cite{cao2023masactrl} proposed the mutual-self attention and copied the K and V in the later diffusion process to adapt to the layout change but deteriorates object and style change. Similarly, InfEdit~\cite{xu2023inversion} combines cross-attention and mutual-self attention to mitigate the deficiency of rigid and non-rigid editing. However, this may degrade the fidelity of edited images. The third category of the refinement approach~\cite{brack2023sega,koo2024flexiedit} filters out components of the predicted noise corresponding to the non-target regions to preserve the non-target regions. However, these method require careful selecting hyperparameters of the filter. Moreover, these approaches are mainly based on the diffusion U-Net models. To fully leverage the generation prior of flow transformer, it is necessary to develop more flexible invariance control system to reconcile both rigid and non-rigid editing types based on the flow transformer for high fidelity and versatile editing.

\section{Analysis of approximation error of Euler}
\begin{figure}[t]
\centering 
\includegraphics[width=1\linewidth]{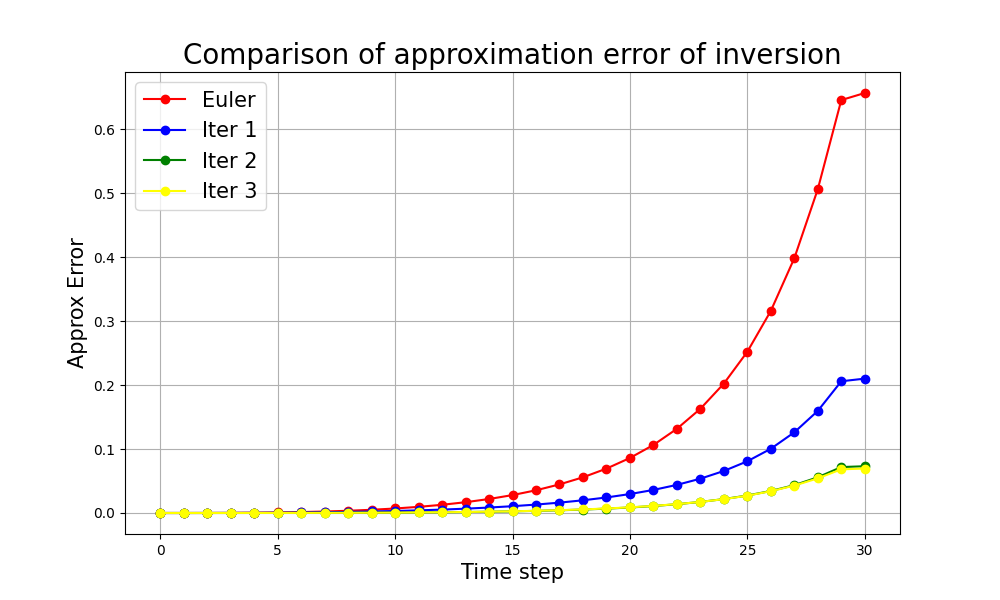}
\caption{\textbf{Comparison of approximation error of inversion with fixed-point iteration at each time step}.}
\label{fig:inv_error_com}
\end{figure}

We compare the approximation error of the plain Euler inversion and the fixed-point iteration in Figure~\ref{fig:inv_error_com}. Concretely, we calculate the MSE difference of two latents at the same time step in inversion $\mathbf{x}_{t}$ and reconstruction $\mathbf{\hat{x}}_{t}$. The result shows that the proposed inversion method can significantly reduce the approximation error at each time step during the reconstruction process. The numerical results also show that increasing the iteration number larger than 3 will lead to marginal improvement on the reconstruction quality. In practice, one iteration can significantly reduce the inversion error.

\section{Versatile image editing with AdaLN invariance control}
\label{app:adaln}
\begin{figure*}[htb]
\centering 
\includegraphics[width=1\linewidth]{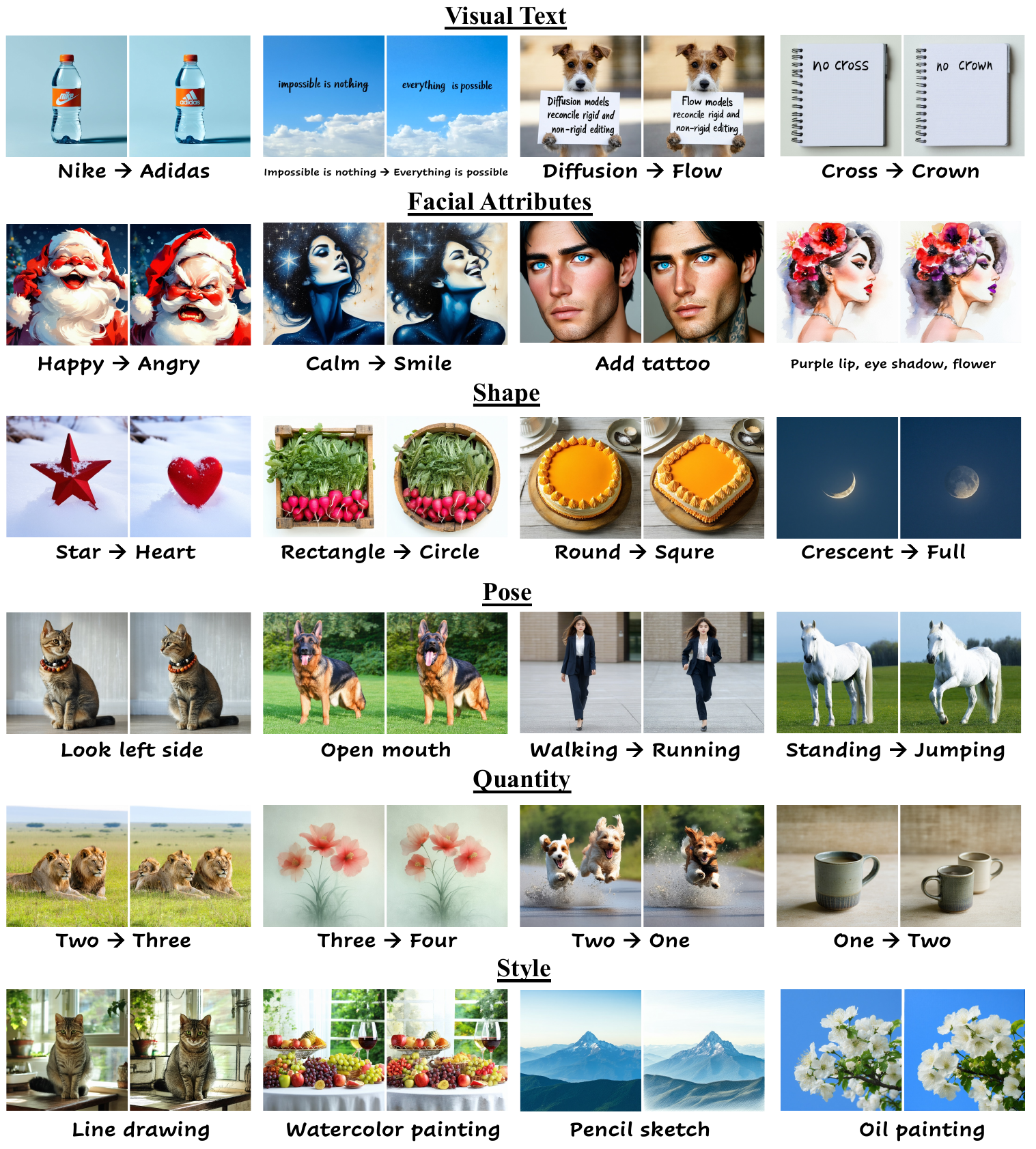}
\caption{\textbf{Qualitative results on versatile editing types Part I}. Zoom in for details.} 
\label{fig:sup_versatile}
\end{figure*}

To validate the adaptability of our method on versatile editing types, especially for non-rigid editing, we categorize the editing results into 12 different editing types. For all editing experiments, we set the timestep $S$ for AdaLN feature replacement as 1 and sampling steps as 30, and summarize the editing results in Figure~\ref{fig:sup_versatile} and Figure~\ref{fig:sup_versatile2}. The results show that our method can reconcile both rigid and non-rigid editing, fully taking advantage of the powerful generation ability of the flow transformer.

Concretely, our method supports the fine control for\textbf{ visual text editing}. The editing results show that our method can manipulate the letter-level visual text such as `cross' to `crown'. This also validates that the feature replacement within AdaLN can discriminately connect the changed text to changed image semantics which is better than the non-discriminative self-attention replacement mechanism. As for the other non-rigid editings, we show results on \textbf{facial attributes, shape, pose, and quantity}. Such editing types change the object structure and layout in the image and demand the invariance control mechanism to have a more precise and flexible ability to manipulate the image semantics. Our method shows accurate and flexible editing results on these non-rigid editing types. For other rigid editing types, our method also shows good generalization performance and can support different levels of geometric changes. For example, our method can distinguish the foreground and background, replace the background with other contents in \textbf{Background change}, and change the whole image style in \textbf{Style}. Our method can also replace small-sized objects, such as `torch to flowers' and `add angels' in \textbf{Object replacement}. In conclusion, the proposed AdaLN invariance control mechanism can support versatile editing types.

\section{Study of attention-based invariance control}
\label{app:self-attn}
\begin{figure*}[t] 
\centering 
\includegraphics[width=1\linewidth]{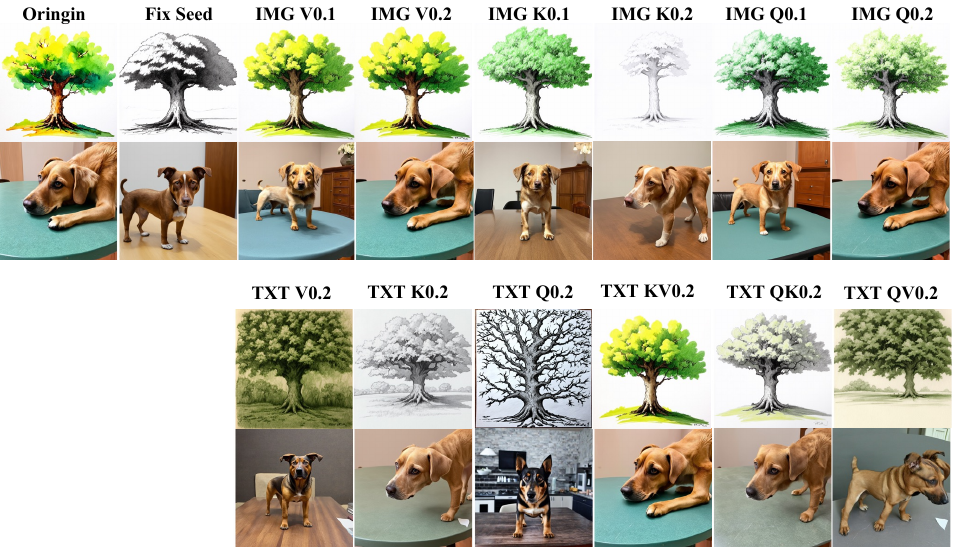}
\caption{\textbf{Investigation of attention-based invariance control in MM-DiT}. Fix seed is the generation results with the same seed but without any invariance control. We inject the Q, K, and V values of text (TXT) and image (IMG) features with different time steps to evaluate the invariance preservation ability. For image features, after the 20\% (IMG0.2) time steps, the injection makes the edited image the same as the original image. For text features, injection Q, K, and V more than 20\% (TXT0.2) time steps do not effectively control the invariance. We further test the combination injection of KV, QK, and QV.}
\label{fig:attn_inv}
\end{figure*}

In comparison to the proposed invariance control based on AdaLN, we also show investigation results based on the attention replacement in Figure~\ref{fig:attn_inv}. Since there is only self-attention in MM-DiT and the text and image features are processed in the attention jointly, we test different strategies of injecting Q, K, and V values of text and image features, instead of the attention map used in P2P~\cite{hertz2022prompt} and MasaCtrl~\cite{cao2023masactrl}. The conclusions are similar to the properties of self-attention in DM models. For image values, injection of the Q, K, and V values from the original image can preserve the contents of the original image but injecting attention values in more steps from the original image will make the edited image overwritten by the original image, which hinders the editing. In the Figure~\ref{fig:attn_inv}, we show that injecting the V or Q values more 20\% time steps will hinder the non-rigid editing `lying to standing', and make the edited image the same as the original image. Injection of K values does not inject the invariant contents of the original image to the target image.

For text values, the results of injecting TXT Q, K, and V do not follow the color and structure pattern. This shows that the individual injection of the Q, K, and V values from the text features cannot effectively preserve the structure of the original image and thus are not enough for invariance control. So, we further test the injection of the combination of KV, QK, and QV values. The results show that the injection of KV features may overwrites the edited image with the original image contents and hinders the non-rigid editing. The injection of QK does not hinder the editing but the edited image fidelity degrades. The injection of QV does not effectively control the invariance, and the edited images are very different from the original images. In conclusion, the attention-based invariance control in MM-DiT has similar properties in U-Net. It is not an effective tool for control invariance for non-rigid editing and overly injecting can influence the editing effect. Inappropriate injection position may degrade the fidelity.

\section{Additional non-rigid editing results}
\label{app:nonrigid}
To further validate the advantages of our method over non-rigid editing, we select 40 images for pose editing in the PIE dataset. The quantitative results are summarized in Table~\ref{tab:nonrigid}. Our method outperforms others in CLIP similarity indicating that our method has the better ability for non-rigid editing. The attention-based methods show a better ability to preserve the contents but cause a lower CLIP score, indicating that their editing abilities are hindered. This is because the images are not edited and remain as the before-images for non-rigid editing types. For example, in the case of `\texttt{bird to X}' in Figure~\ref{fig:real_comp}. This confirms our argument that attention-based invariance control can hinder non-rigid editing ability. Besides, we also recall that in the full PIE benchmark Table~\ref{tab:main}, our method outperforms InfEdit in PSNR, LPIPS, MSE, SSIM, and Distance.

In contrast, PnP takes the second place in non-rigid editing but the preservation is much worse than ours. The PSNR, MSE, and SSIM are obviously worse than our method. In conclusion, our method can do good non-rigid editing while maintaining a reasonable ability to preserve the background contents.

\begin{table}[h]
\small
\centering
\tabcolsep=0.1cm
\caption{\textbf{Quantitative comparisons in non-rigid image editing.} Evaluated using the PIE benchmark. Different metrics are scaled.}
\resizebox{1\linewidth}{!}{
\begin{tabular}{@{} l S S S S S S S @{}}
\toprule
& \multicolumn{1}{c}{Structure} & \multicolumn{4}{c}{Background Preservation} & \multicolumn{2}{c}{CLIP Similarity} \\
\cmidrule(lr){2-2} \cmidrule(lr){3-6} \cmidrule(lr){7-8}
\scriptsize \textbf{Method} & \scriptsize{$\textbf{Distance}$ $\downarrow$} & \scriptsize{$\textbf{PSNR}$ $\uparrow$} & \scriptsize{$\textbf{LPIPS}$ $\downarrow$} & \scriptsize{$\textbf{MSE}$ $\downarrow$} & \scriptsize{$\textbf{SSIM}$ $\uparrow$} & \scriptsize{$\textbf{Whole}$ $\uparrow$} & \scriptsize{$\textbf{Edited}$ $\uparrow$} \\
\midrule
PnP  & 20.63 & 22.76 & 116.36 & 79.66 & 77.58 &\underline{26.64} &\underline{22.82} \\
P2P  &\underline{9.42} &\underline{26.63} &\underline{59.28} &\underline{32.94} &83.64 &26.56 &22.43 \\
MasaCtrl  & 20.53 & 22.47 & 91.84 &  89.11 & 79.79 & 26.59 & 22.38 \\
InsP2P  &53.38 &20.82 &165.33 &243.23 &72.18 &23.35 &20.13 \\
InfEdit  & \textbf{6.12} & \textbf{28.35} & \textbf{43.36} &  \textbf{23.64} & \underline{85.35} & 25.98 & 22.18 \\
Ours & 20.39 & 24.02 & 103.73 &53.70 & \textbf{87.47} & \textbf{26.93} & \textbf{22.87} \\
\bottomrule
\end{tabular}
}
\label{tab:nonrigid}
\end{table}

\begin{figure*}[t]
\centering 
\includegraphics[width=1\linewidth]{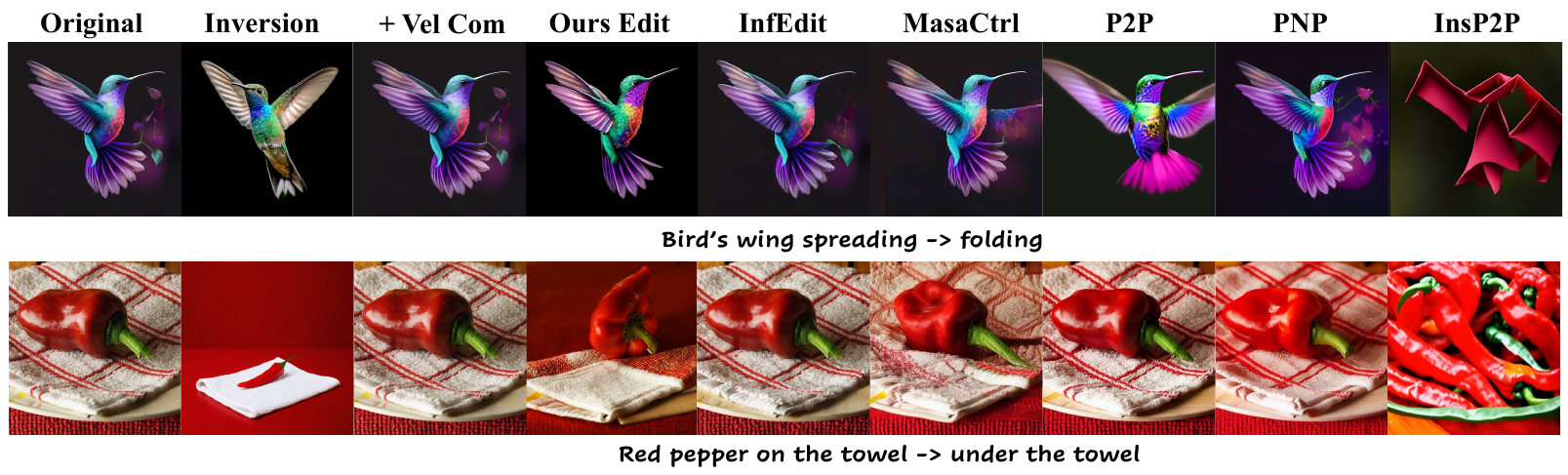}
\caption{\textbf{Failure case study}. We present two failure cases whose inversions are far away from the original image. The inversions presented are processed with 1 fixed-point iteration. Even though the velocity compensation can recover the original image, the editing still fails.}
\label{fig:failure}
\end{figure*}

\section{Failure case study}
We present the failure case study to better understand the limitations of our framework. So far, we already demonstrated the superior advantages of the invariance control based on the AdaLN. However, as discussed in the Limitation section, if the real image is not within the prior distribution of the model, the inversion is far from the authentic generation process and is quite different from the original image. Thus, the text-to-image alignment based on the reconstructed inversion trajectory is mismatched. In this case, changing the text prompt may also cause changes in non-target regions. We show the case in Figure~\ref{fig:failure}, and the result shows that if the inversion (without the velocity compensation) seriously deviates from the original image, the output image also cannot be properly edited even if the image can be fully recovered with the velocity compensation. We also show the results of other methods, and none of the methods successfully made the right edit since this image may not be within the domain of the model.

\begin{figure*}[htb] 
\centering 
\includegraphics[width=1\linewidth]{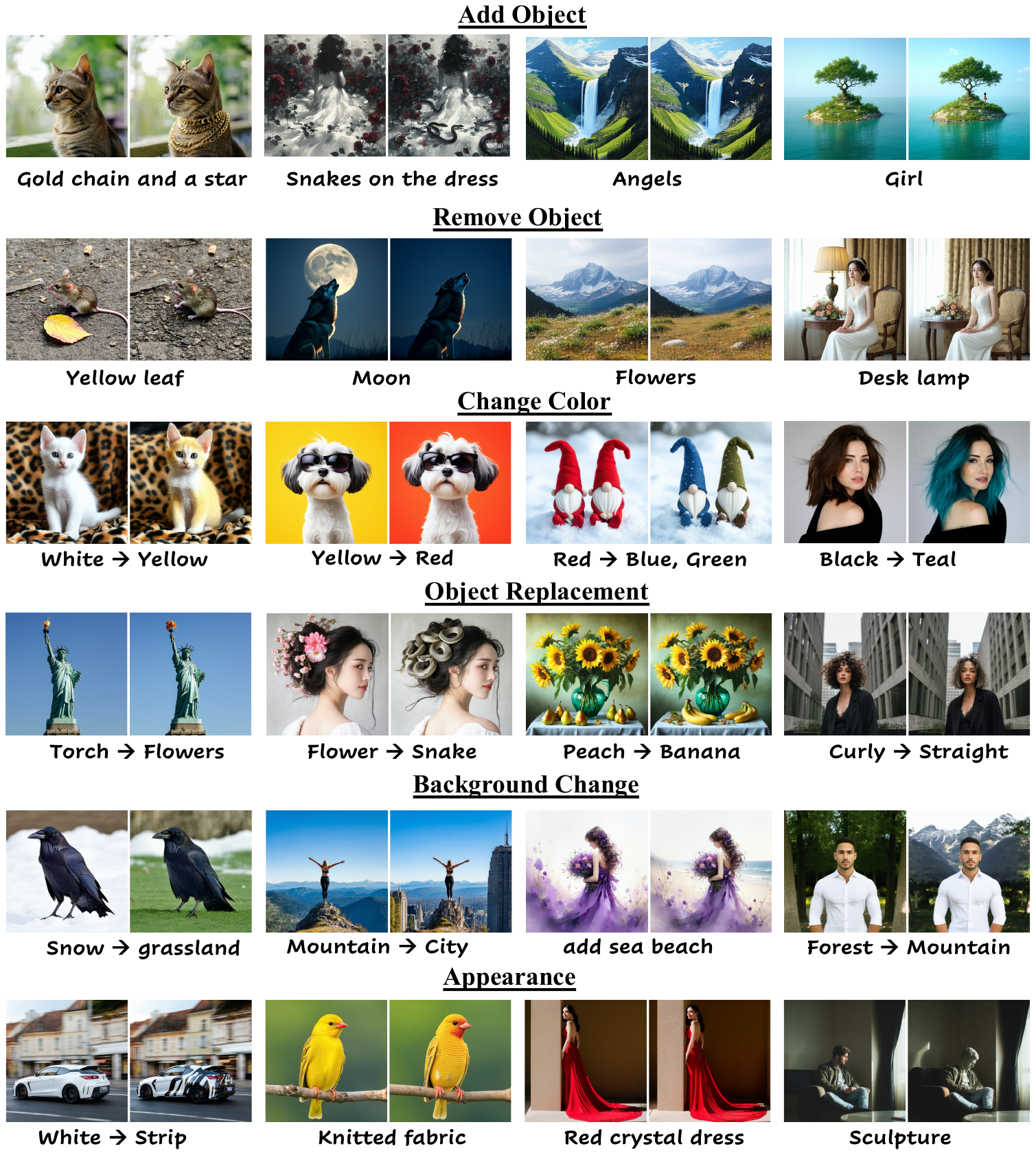}
\caption{\textbf{Qualitative results on versatile editing types Part II}. Zoom in for details.} 
\label{fig:sup_versatile2}
\end{figure*}

\begin{figure*}[htb] 
\centering 
\includegraphics[width=1\linewidth]{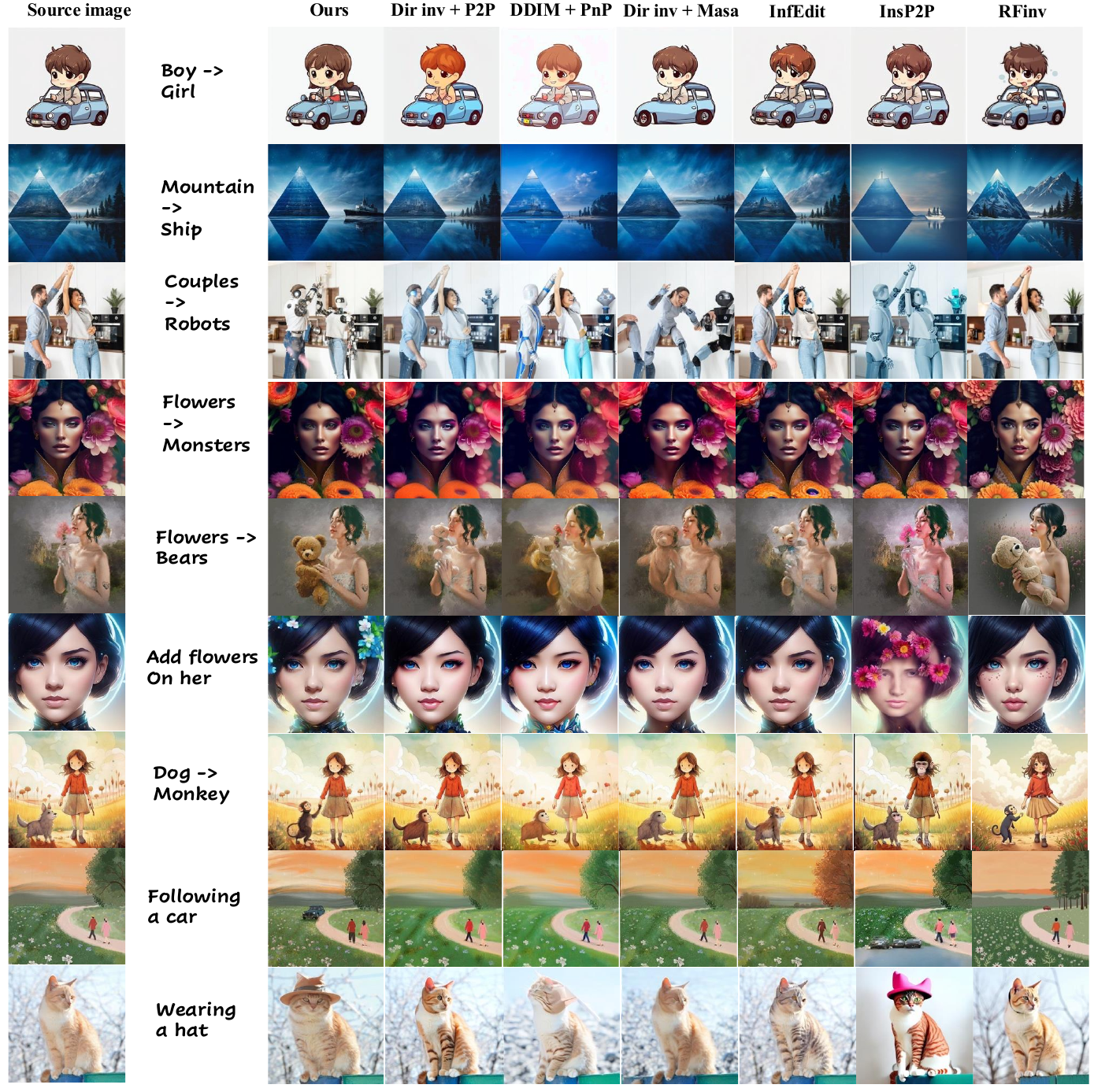}
\caption{\textbf{Additional qualitative results on PIE benchmark}. Zoom in for details.} 
\label{fig:sup_versatile2}
\end{figure*}

\end{document}